\documentclass[sigconf]{acmart}
\usepackage{algorithm}
\usepackage{algorithmic}
\usepackage{amsmath}
\usepackage{amsfonts}
\usepackage{booktabs}
\usepackage{nicefrac}       
\usepackage{microtype}      
\usepackage{xcolor}         
\usepackage{multirow}
\usepackage{graphicx}
\usepackage{makecell}
\usepackage{float}
\usepackage[stable]{footmisc}
\usepackage{newfloat}
\usepackage{listings}

\AtBeginDocument{%
  }

\setcopyright{acmlicensed}
\copyrightyear{2018}
\acmYear{2018}
\acmDOI{XXXXXXX.XXXXXXX}
\acmConference[Conference acronym 'XX]{Make sure to enter the correct
  conference title from your rights confirmation email}{June 03--05,
  2018}{Woodstock, NY}
\acmISBN{978-1-4503-XXXX-X/2018/06}

\begin{document}

\title{m\textsuperscript{3}BERT: A Modern, Multi-lingual, Matryoshka  Bidirectional Encoder}
  
\author{Yaoxiang Wang}
\authornote{This work is done during their internships at Microsoft.}
\email{wangyaoxiang@stu.xmu.edu.cn}
\affiliation{%
  \institution{Xiamen University}
  \city{Xiamen}
  \country{China}
}

\author{Simiao Zuo}
\email{simiaozuo@microsoft.com}
\affiliation{%
  \institution{Microsoft}
  \city{Redmond}
  \country{USA}}

\author{Qingguo Hu}
\email{huqingguo@stu.xmu.edu.cn}
\authornotemark[1]
\affiliation{%
  \institution{Xiamen University}
  \city{Xiamen}
  \country{China}
}

\author{Yucheng Ding}
\email{yc.ding@sjtu.edu.cn}
\authornotemark[1]
\affiliation{%
  \institution{Shanghai Jiao Tong University}
  \city{Shanghai}
  \country{China}
}

\author{Yeyun Gong}
\email{yegong@microsoft.com}
\authornote{Corresponding author.}
\affiliation{%
  \institution{Microsoft}
  \city{Beijing}
  \country{China}}

\author{Jian Jiao}
\email{Jian.Jiao@microsoft.com}
\affiliation{%
  \institution{Microsoft}
  \city{Redmond}
  \country{USA}}

\author{Jinsong Su}
\email{jssu@xmu.edu.cn}
\authornotemark[2]
\affiliation{%
  \institution{Xiamen University}
  \city{Xiamen}
  \country{China}
}
\renewcommand{\shortauthors}{Trovato et al.}

\begin{abstract}
Embedding models are pivotal in industrial information retrieval systems like search and advertising. However, existing pretrained models often exhibit fixed architectures and embedding dimensionalities, posing significant challenges when adapting them to diverse deployment scenarios with varying business-driven constraints. A common practice involves fine-tuning with partial parameter initialization from larger pretrained models for resource-constrained tasks. This method is often suboptimal as the misalignment between pretraining and downstream usage prevents full realization of pretraining benefits. To address this limitation, we introduce $\text{m}^3\text{BERT}$: a \textbf{M}odern, \textbf{M}ulti-lingual, \textbf{M}atryoshka Bidirectional Encoder, which features a novel pretraining strategy that jointly optimizes representations across both transformer layers and multiple embedding dimensions. This enables a single model to be tailored to varied resource and accuracy targets while maintaining consistency with pretraining. Incorporating recent architectural improvements, $\text{m}^3\text{BERT}$ uses a three-stage pretraining: monolingual pretraining, multilingual adaptation to serve diverse user bases, and crucial continual pretraining on a massive web domain corpus to enhance utility in commercial retrieval. $\text{m}^3\text{BERT}$ significantly outperforms state-of-the-art embedding models in \textsc{Bing-Click}, a large-scale industrial retrieval dataset, showcasing its practical versatility as an efficient foundation for resource-aware industrial retrieval systems. Further experiments on public datasets also confirm the general effectiveness of our multigranular Matryoshka pretraining strategy.
\end{abstract}

 



\maketitle

\section{Introduction}

The embedding model is a fundamental component in information retrieval, especially encoder-only transformer architectures~\citep{embedsurvey}, which are widely used in commercial applications such as search engines, advertising platforms, retrieval-augmented generation~\citep{xiang2025use}, and agent systems~\citep{gao2025tool}. In these domains, pretraining on large-scale datasets followed by supervised fine-tuning with domain-specific data has proven to be an effective approach for building robust embedding models.

\begin{figure*}[ht]
    \centering
    \includegraphics[width=1.0\textwidth]{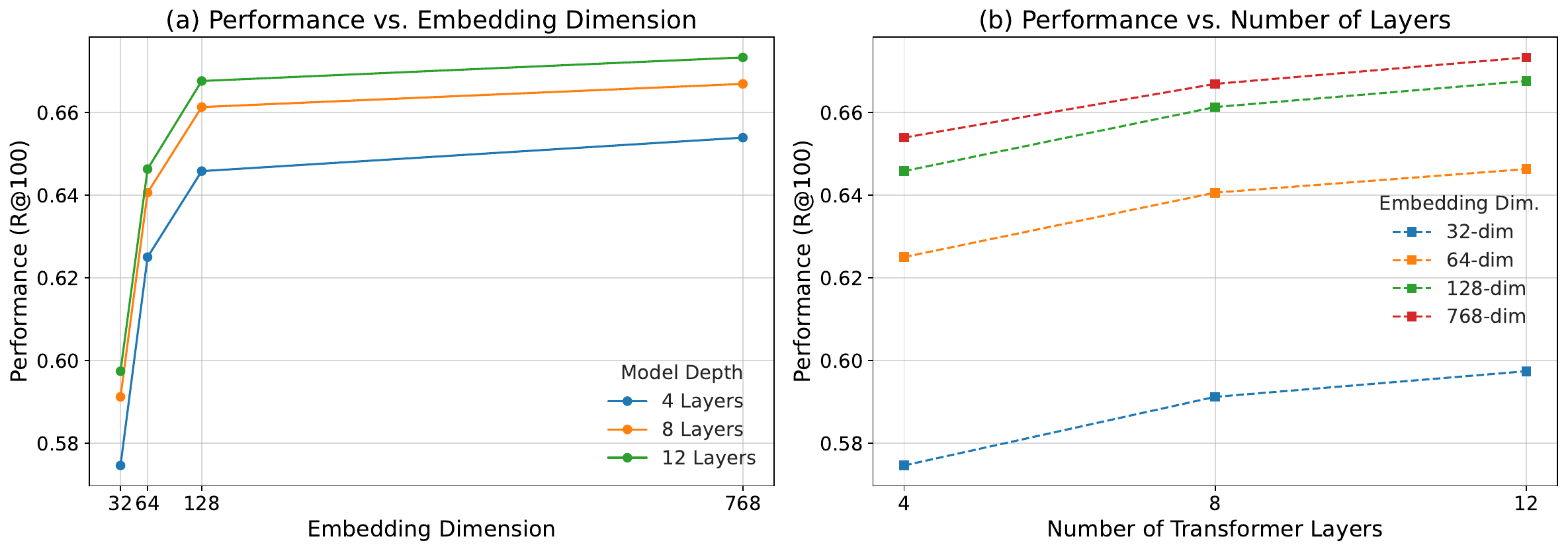}
    \caption{Illustrative curves showing the diminishing returns of retrieval performance (Recall@100) with increasing (a) embedding dimension and (b) number of transformer layers for an mBERT model on \textsc{Bing-Click}. While performance generally improves, the gains diminish significantly at higher dimensions/depths, whereas computational costs (e.g., forward latency, embedding memory) often scale linearly.}
    \label{fig:tradeoff_curves}
\end{figure*}

Existing pretrained embedding models~\citep{bert,roberta,e5,modernbert} provide powerful general-purpose embeddings and often incorporate training strategies and architectural innovations to enhance performance. However, a significant limitation of these models lies in their rigid design: both the model size and the dimension of the generated embeddings are fixed. This rigidity makes it challenging to meet the diverse requirements of downstream tasks, which may demand varying trade-offs between retrieval performance, latency, and computational resources. Figure~\ref{fig:tradeoff_curves} illustrates this challenge: while increasing embedding dimension or model depth (number of layers) generally improves retrieval performance, the gains often diminish significantly at higher values.
For instance, doubling the embedding dimension from 64 to 128 might yield a substantial performance boost, but further increasing it to 768 may offer only marginal improvements while linearly increasing retrieval latency and storage costs.

This characteristic of diminishing return underscores the inefficiency of a one-size-fits-all approach in embedding model design. Tasks with strict latency constraints or those deployed on resource-constrained hardware may thus require smaller models and lower-dimensional embeddings to operate efficiently, even if it means a slight compromise on performance. To obtain these smaller variants from an existing large pre-trained model, a common practice is to construct a smaller architecture and initialize its parameters by copying a corresponding weights from the large model.  However, this approach of partial parameter initialization can fail to fully leverage the benefits of pre-training due to the misalignment between pretraining and downstream fine-tuning, thereby limiting the performance gains.

To address this challenge, we draw inspiration from Matryoshka Representation Learning (MRL)~\citep{mrl}, which enables flexibility in embedding representations by training them at multiple dimensions.
Extending this concept, we propose a novel pre-training strategy that incorporates multigranular embedding representations across both transformer layers and embedding dimensions. By jointly optimizing diverse embeddings along these two axes during pretraining, our approach equips the model with flexible and robust representations, making it a stronger foundation for fine-tuning across a broad range of downstream tasks requiring different model sizes and embedding dimensions.

Building upon this strategy, we present $\text{m}^3\text{BERT}$, a \textbf{M}odern, \textbf{M}ulti-lingual, \textbf{M}atryoshka Bidirectional Encoder, designed for flexible adaptation to varying deployment scenarios. While current widely used embedding models ~\cite{e5,multi2024m3, solon_embeddings}predominantly adopt the traditional BERT architecture, recent advancements in language model design ~\citep{llama, jung2010mistral, team2403gemma} have indicated potential sub-optimality in certain aspects of this structure. Consequently, we incorporate recent architectural improvements from large language models into our $\text{m}^3\text{BERT}$. Our pre-training process comprises three distinct stages: monolingual pretraining, multilingual pretraining, and continual pretraining on a large-scale web domain corpus. This three-stage approach is specifically designed to enhance the model's performance in real-world retrieval tasks.
 
Extensive experiments demonstrate the effectiveness and robustness of our approach. On our collected large-scale industrial dataset BINGCLICK, $\text{m}^3\text{BERT}$ consistently outperforms current state-of-the-art embedding models across various model sizes and embedding dimensions. To further validate the generalizability of our matryoshka pre-training strategy, we conduct evaluations on multiple public datasets, showing that its benefits extend beyond our specific industrial use case. Additionally, we explore a novel self-distillation technique, termed Matryoshka Distillation, which leverages the nested structure of our model to further enhance the performance of smaller embeddings during the pre-training phase.

The effectiveness of $\text{m}^3\text{BERT}$ is validated by its large-scale deployment at Bing Search. Since June 2025, 
$\text{m}^3\text{BERT}$ has been a core component of the production query-keyword selection workflow, consistently handling over 25,000 queries per second (QPS). This deployment, which contributes an annualized revenue impact of approximately USD 50 million, demonstrates the tangible business value and engineering success of our model design.
 
Our main contributions can be summarized as follows:

\begin{itemize}
    \item We introduce the Matryoshka Representation Learning paradigm into the pretraining of embedding models and extend it to encompass multigranular representations across both transformer layers and embedding dimensions. This innovation enables a single pretrained model to effectively align with the varying deployment scenarios.
    \item We pretrain $\text{m}^3\text{BERT}$, a novel model that incorporates modern architectural advancements. We leverage a massive, multi-stage pre-training process, including a large-scale web domain corpus, to specifically tailor the model for superior performance in industrial retrieval applications.
    \item Our model not only outperforms state-of-the-art baselines in extensive offline experiments but has also been proven under live traffic at Bing, delivering substantial business impact with an annualized revenue of USD 50 million.
\end{itemize}

\section{$\text{m}^3\text{BERT}$}

\begin{figure*}[ht]
    \centering
    \includegraphics[width=1.0\textwidth]{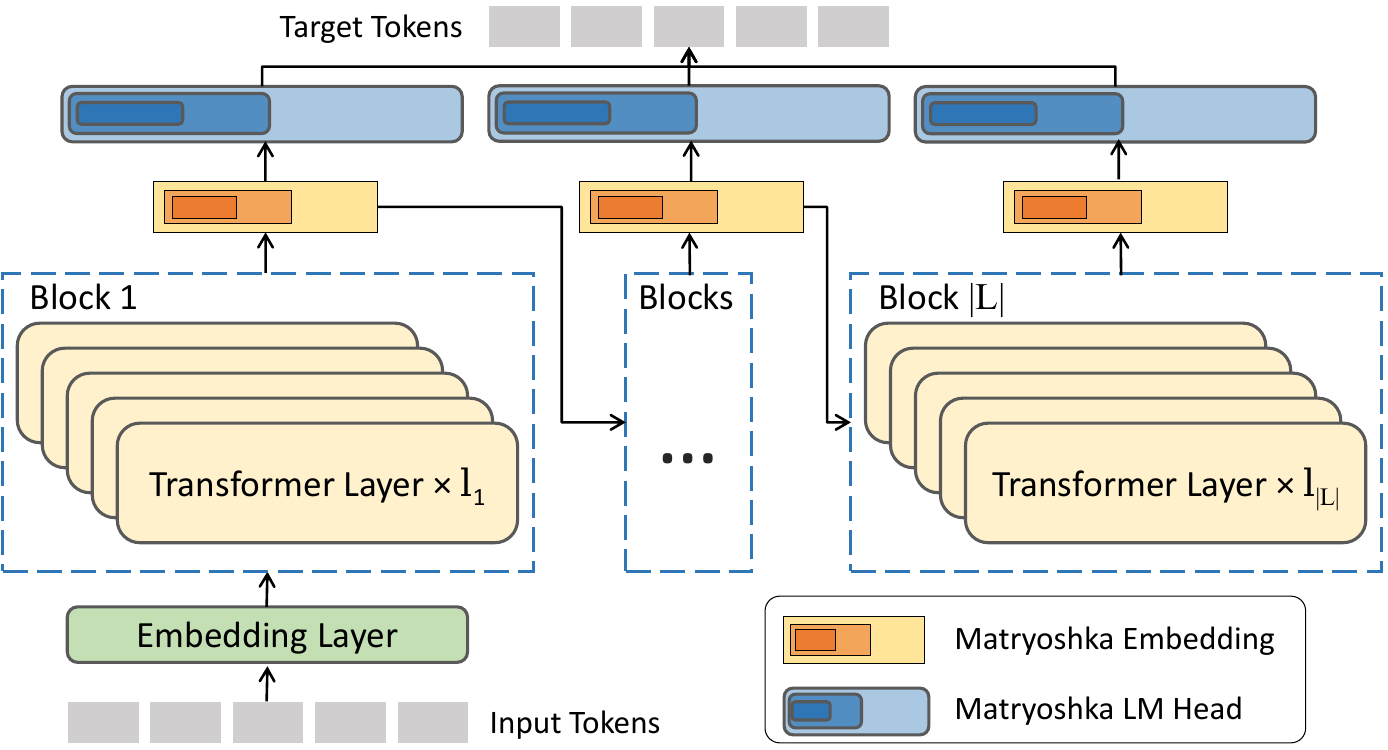}
    \caption{Overview of the matryoshka model structure using masked language modeling (MLM) as the training objective. The model simultaneously optimizes embeddings across multiple transformer layers and multiple sub-dimensions.}
    \label{fig:matryoshka_structure}
\end{figure*}

\subsection{Matryoshka Pretraining}
\label{sec:method_mrl}
To enable flexible and efficient embedding across various downstream tasks, we propose a multigranular embedding pretraining strategy. Unlike traditional pre-training approaches that rely solely on the final layer and full-dimensional embeddings, our method jointly optimizes embeddings across multiple transformer layers and multiple sub-dimensions.

Formally, let a transformer encoder have $N$ total layers, producing hidden states $\{h^k\}_{k=1}^N$. Each full hidden state from a layer $k$ is $h^k \in \mathbb{R}^{s \times M}$, where $s$ is the sequence length, and $M$ is the full embedding dimension. To accommodate diverse deployment scenarios, we predefine a set of selected layer indices $L \subseteq \{1, \ldots, N\}$ and a set of target sub-dimensions $D \subseteq \{1, \ldots, M\}$. For each selected layer $l_i \in L$ and each sub-dimension $d_j \in D$, we extract the truncated embedding from the output of layer $l_i$:
\[
h^{l_i}_{[:d_j]} \in \mathbb{R}^{s \times d_j}
\]
where $h^{l_i}_{[:d_j]}$ denotes taking the first $d_j$ dimensions of $h^{l_i}$ (the full $M$-dimensional hidden state output by layer $l_i$). This process results in $|L| \times |D|$ distinct embedding representations that are jointly optimized during training.

For masked language modeling (MLM) training, we reuse a shared MLM head. The MLM head consists of a projection matrix $W \in \mathbb{R}^{M \times V}$ and a bias $b \in \mathbb{R}^{V}$, where $V$ is the vocabulary size. When using a truncated embedding $h^{l_i}_{[:d_j]}$, we utilize the corresponding submatrix $W_{[:d_j,:]} \in \mathbb{R}^{d_j \times V}$ to project the embedding into the vocabulary space:
\[
\hat{y}^{l_i, d_j} = \text{Softmax}\left(h^{l_i}_{[:d_j]} W_{[:d_j,:]} + b\right)
\]

The total loss is computed by aggregating the MLM losses across all selected (layer, sub-dimension) pairs:
\[
\mathcal{L}_{\text{total}} = \sum_{l_i \in L} \sum_{d_j \in D} \mathcal{L}_{\text{MLM}}(\hat{y}^{l_i, d_j}, y)
\]
where $y$ denotes the ground-truth tokens at masked positions.

By optimizing $\mathcal{L}_{\text{total}}$, the model learns robust and flexible representations that are consistent and predictive across multiple granularities of layers and embedding dimensions. This design ensures that practitioners can dynamically select different model depths and embedding widths to balance the trade-offs between latency, memory, and performance.

\subsection{Modern Architecture}
Building upon the original BERT architecture, we incorporate several recent advancements from large language models (LLMs) to enhance training and inference efficiency, improve training stability, and boost overall model performance.

\paragraph{Activation Function.} 
We replace the GeLU activation of BERT with SwiGLU~\citep{swiglu}, which has demonstrated better performance in transformer-based models~\citep{modernbert, llama, olmo}. The SwiGLU activation introduces a gated mechanism, providing stronger non-linearity and facilitating more effective feature learning.

\paragraph{Normalization.} 
Following recent trends in LLM~\citep{llama, qwen}, we adopt the root mean square layer normalization (RMSNorm)~\citep{rmsnorm} instead of the standard LayerNorm. Additionally, we transition from post-norm configuration to a pre-norm design, which stabilizes training and improves gradient flow in deep transformers~\citep{prenorm}.

\paragraph{Bias Terms and Dropout.} 
Recent studies~\citep{modernbert, llama} indicate that bias terms contribute little to model performance while introducing unnecessary computational overhead. We remove the bias terms in both the self-attention and feed-forward layers. Furthermore, we eliminate dropout applied to the hidden states between transformer layers, as its removal has been shown to maintain performance while simplifying the model and accelerating training.

\paragraph{Flash Attention.} 
We replace the standard attention computation with FlashAttention~\citep{flashattention, flashattention2}, an efficient algorithm that leverages tiling and memory-efficient strategies to significantly accelerate attention operations without compromising numerical precision.

By integrating these architectural refinements, our model inherits the proven strengths of BERT while benefiting from the efficiency and scalability improvements pioneered in recent LLMs.

\subsection{Three-Stage Pretraining}
To equip the model with broad linguistic understanding, cross-lingual capabilities, and domain-specific knowledge, we adopt a three-stage pretraining strategy.

\paragraph{Stage 1: Monolingual Pretraining.}
We first conduct pretraining on a large English corpus consisting of 100 billion tokens from Nemotron-CC~\citep{su2024nemotron} using the masked language modeling (MLM) objective. Although previous work~\citep{wettig2022should} suggests that increasing the masking rate to $0.3$ can benefit the training of larger models, we observe that in our setting---which emphasizes learning from reduced layers and sub-dimensional embeddings---a high masking rate leads to unstable training and convergence issues. Therefore, we adopt a standard masking rate of $0.15$ to balance learning difficulty and training stability.

\paragraph{Stage 2: Multilingual Pretraining.}
To extend the model’s capabilities to multilingual scenarios, we continue pretraining on a corpus constructed from the Wikipedia dumps of the top 100 languages, totaling approximately 20 billion tokens. Following the general practice introduced by multilingual BERT~\citep{bert}, we apply an exponential smoothing strategy to balance the contribution of different languages. Specifically, if $P(L)$ denotes the original data proportion for language $L$, the sampling probability is adjusted as $P'(L) \propto P(L)^S$ with a smoothing factor $S=0.7$, followed by normalization. This approach mitigates the imbalance between high-resource and low-resource languages by relatively downsampling dominant languages like English and upsampling underrepresented languages.

\paragraph{Stage 3: Web Domain Pretraining.}
To specialize the model for ads and web search-related applications, we adapt our multigranular pretraining strategy using an infinite contrastive learning objective (Inf-CL)~\citep{infcl}  on in-domain data. Inf-CL facilitates training with exceptionally large batch sizes by utilizing a tile-based computation strategy. 
This approach partitions the contrastive loss calculation, thereby avoiding the full materialization of the similarity matrix and overcoming typical GPU memory constraints associated with scaling contrastive learning.
The training corpus consists of 10 billion query-document (Q-Doc) pairs collected from one month of user-ads click logs. Each pair includes ad features (e.g., keyword, ad titles and ad descriptions), and URL features (e.g., landing page titles and contents). We apply extensive data cleaning, including Q-Doc deduplication and query-level sampling caps. Pretraining continues from the Stage 2 checkpoint under the same multigranular learning settings, with batch sizes scaled up to one million samples. The model is trained for 2 trillion tokens over the corpus.

\paragraph{Training Details.} For our model size, we adopt the same configuration as mBERT-base\footnote{\url{https://huggingface.co/google-bert/bert-base-multilingual-uncased}} and mE5-base\footnote{\url{https://huggingface.co/intfloat/multilingual-e5-base}}, utilizing 12 transformer layers and a hidden size of 768.
In Stage 1 and Stage 2, we adopt masked language modeling (MLM) as the training objective. For the multigranular optimization, we select a layer set $L=\{4, 8, 12\}$ and a dimension set $D=\{32, 64, 128, 768\}$. These granularities are specifically chosen to align with the hierarchical deployment tiers and hardware constraints in Bing's production infrastructure, where different retrieval stages must adhere to latency budgets and memory limits. The learning rate was set to $1 \times 10^{-4}$ with a batch size of 128K sequences, each with a maximum length of 512 tokens. In Stage 3, we switch to the Infinite contrastive learning (Inf-CL) objective to better align with the web domain data, increasing the batch size to 1 million samples and setting the learning rate to $2 \times 10^{-4}$. For this stage, considering computational efficiency and practical application needs, the multigranular optimization utilizes a layer set $L=\{4, 12\}$ and a dimension set $D=\{32, 64, 128\}$. Across all stages, we employ the AdamW optimizer~\citep{adamw} with $\beta_1=0.9$, $\beta_2=0.999$, a weight decay of $0.01$, and a cosine learning rate decay schedule with 10,000 warm-up steps. To enhance computational efficiency, bfloat16 mixed-precision training was utilized. The entire pretraining process was conducted on a cluster of NVIDIA A100 GPUs, accumulating approximately 40,000 GPU hours.

\section{Experiments}
This section presents a comprehensive empirical evaluation of our proposed framework. 
In Section~\ref{sec:setup}, we detail the experimental setup, including our large-scale industrial dataset \textsc{Bing-Click}. 
Section~\ref{sec:main} showcases the main results on \textsc{Bing-Click}, demonstrating that $\text{m}^3\text{BERT}$ achieves state-of-the-art performance in this industrial retrieval scenario. To validate the core of our method, Section~\ref{sec:matryoshka_pretrain_impact} isolates the impact of the multigranular matryoshka pretraining strategy by comparing it against standard pretraining across multiple datasets. Finally, Section~\ref{sec:deployment} discusses the successful deployment of our model, highlighting its operational scale and significant business impact.

\begin{table*}[ht]
\centering
\caption{Main retrieval performance (Recall@100 and Recall@1000) on the \textsc{Bing-Click} test set. Results are grouped by model configuration (Lite or Full layers). Best performing figures for each metric and configuration are in \textbf{bold}. $\text{m}^3\text{BERT-s3}$ is not evaluated at 768 dim due to its matryoshka pretraining focus up to 128 dim.}
\label{tab:main_results}
\begin{tabular}{@{}lcccccccc@{}}
\toprule
\multirow{2.5}{*}{Model} & \multicolumn{2}{c}{Dim=32} & \multicolumn{2}{c}{Dim=64} & \multicolumn{2}{c}{Dim=128} & \multicolumn{2}{c}{Dim=768} \\
\cmidrule(lr){2-3} \cmidrule(lr){4-5} \cmidrule(lr){6-7} \cmidrule(lr){8-9}
                       & R@100 & R@1000 & R@100 & R@1000 & R@100 & R@1000 & R@100 & R@1000 \\
\midrule
\multicolumn{9}{@{}l}{\textit{Lite (1/3 layers)}} \\
\quad mBERT                      &  57.46 &  80.78 &  62.50 &  84.67 &  64.58 &  86.27 &  65.39 &  86.99 \\
\quad mE5                        &  57.18 &  80.63 &  62.16 &  84.40 &  64.40 &  86.14 &  65.39 &  86.90 \\
\quad ModernBERT                 &  57.10 &  80.55 &  62.43 &  84.59 &  64.59 &  86.23 &  65.34 &  86.87 \\
\quad $\text{m}^3\text{BERT-s2}$ &  58.34 &  81.67 &  63.43 &  85.49 &  65.47 &  87.02 & \textbf{66.33} & \textbf{87.66} \\
\quad $\text{m}^3\text{BERT-s3}$  & \textbf{62.06} & \textbf{84.52} & \textbf{66.47} & \textbf{87.67} & \textbf{67.72} & \textbf{88.51} & -- & -- \\
\midrule
\multicolumn{9}{@{}l}{\textit{Full}} \\
\quad mBERT                      &  59.74 &  82.83 &  64.63 &  86.28 &  66.76 &  87.69 &  67.33 &  88.12 \\
\quad mE5                        &  59.31 &  82.24 &  64.32 &  85.98 &  66.85 &  87.39 &  67.97 &  88.49 \\
\quad ModernBERT                 &  59.44 &  82.65 &  64.62 &  86.27 &  66.75 &  87.71 &  67.50 &  88.25 \\
\quad $\text{m}^3\text{BERT-s2}$ &  60.25 &  83.29 &  65.44 &  87.00 &  67.34 &  88.37 & \textbf{68.32} & \textbf{89.00} \\
\quad $\text{m}^3\text{BERT-s3}$  & \textbf{63.30} & \textbf{85.50} & \textbf{67.60} & \textbf{88.35} & \textbf{68.94} & \textbf{89.23} & -- & -- \\
\bottomrule
\end{tabular}%
\end{table*}

\subsection{Setup}
\label{sec:setup}
\paragraph{Datasets}
To evaluate the effectiveness of $\text{m}^3\text{BERT}$ in real-world industrial retrieval scenarios, we collected and curated a large-scale multilingual industrial text retrieval dataset, which we name \textsc{Bing-Click}. This dataset comprises a substantial volume of user queries and their corresponding clicked webpages, gathered from the Bing search engine across diverse languages and regions worldwide. Specifically, a page is considered a positive instance for a given user query if the user clicked on it and subsequently remained for ten seconds, indicating relevance. We amassed six months of such interaction data, resulting in 100 million query-page pairs after being filtered for training.
For evaluation, we constructed a test set using the same methodology in another three months, ensuring that the click timestamps for the test data chronologically followed those of the training data to prevent data leakage and simulate a realistic deployment scenario. The test set is designed to emulate a large-scale retrieval system, consisting of a candidate pool of 10 million unique documents and 1 million positive query-document pairs. This massive scale and the temporal separation provide a high-fidelity proxy for the model's robustness against evolving user trends in an online environment. To further validate the generalizability of matryoshka Pretraining beyond the industrial setting, we also conduct experiments on three public benchmarks: \textsc{MS MARCO} document ranking~\cite{msmarco}, \textsc{Natural Questions}~\cite{naturalquestions}, and \textsc{TREC-COVID}~\cite{trec}. More training details can be found in Appendix~A.1. 

\paragraph{Models}
Our proposed models are evaluated in two configurations based on their pretraining:
\begin{itemize}
\item \textbf{$\text{m}^3\text{BERT-s2}$}: This model corresponds to the checkpoint obtained after Stage 1 (English pretraining) and Stage 2 (multilingual adaptation), primarily trained with the multigranular MLM objective.
\item \textbf{$\text{m}^3\text{BERT-s3}$}: This model is the result of Stage 3, which is continually pretrained on the large-scale web domain corpus using the Inf-CL objective, also incorporating the multigranular learning strategy for layers and dimensions.
\end{itemize}

We compare $\text{m}^3\text{BERT}$ against commonly used baseline embedding models of similar parameter scales to ensure a fair and meaningful comparison. These include:
\begin{itemize}
\item \textbf{mBERT}\citep{bert}: We use bert-base-multilingual, a standard multilingual BERT model pretrained with masked language modeling and next sentence prediction objectives.
\item \textbf{mE5}\citep{e5}: We use multilingual-e5-base, a multilingual version of the E5 model designed for dense retrieval.
\item \textbf{ModernBERT}~\citep{modernbert}: We use ModernBERT-base, incorporating modern architectural improvements, pretrained with MLM and a contrastive learning objective.
\end{itemize}
We do not include significantly larger models such as \texttt{bge-m3}~\cite{multi2024m3}. To the best of our knowledge, embedding models used in some large-scale commercial scenarios are in the “lite” or base-size.

To assess performance under stricter resource constraints and to leverage the matryoshka capabilities, we evaluate "lite" versions of all models, which utilize the first 1/3 of their total transformer layers. For embedding dimension, we report results for 32, 64, 128, and 768 dimensions. A notable exception is  $\text{m}^3\text{BERT-s3}$
; due to its multigranular pretraining in Stage 3 being optimized for matryoshka embeddings up to 128 dimensions, we evaluate it for 32, 64, and 128 dimensions.

\paragraph{Supervised Finetuning}
All models are supervised fine-tuned (SFT) using a contrastive learning objective with in-batch negatives. 
Given a query $q$, its corresponding positive document $d^+$, and a set of $N-1$ negative documents $\{d^-_i\}_{i=1}^{N-1}$ from the same batch, we first define the scaled similarity score as $s(q, d) = \text{sim}(E(q), E(d)) / \tau$. Here, $E(\cdot)$ is the embedding function (the model), $\text{sim}(\cdot, \cdot)$ is the cosine similarity, and the temperature $\tau$ is set to $0.05$. The contrastive loss for the query is then formulated as:
\begin{equation}
\mathcal{L}_{\text{SFT}} = -\log \frac{\exp(s(q, d^+))}{\exp(s(q, d^+)) + \sum_{i=1}^{N-1} \exp(s(q, d^-_i))}
\label{eq:sft_loss}
\end{equation}

For \textsc{Bing-Click}, we use a batch size of 1024 and a learning rate of $5 \times 10^{-4}$. For \textsc{MS MARCO}, \textsc{Natural Questions}, and \textsc{TREC-COVID}, we use a batch size of 32 and a learning rate of $2 \times 10^{-5}$. The models are trained on 8 NVIDIA A100 GPUs using the AdamW optimizer~\citep{adamw} with \texttt{fp16} mixed-precision training. Results are averaged over three runs. More training details can be found in Appendix~A.1.

\subsection{Main Results}
\label{sec:main}

We evaluate Recall@100 and Recall@1000 of all models on the \textsc{Bing-Click} test set. The comprehensive results, comparing $\text{m}^3\text{BERT-s2}$ and $\text{m}^3\text{BERT-s3}$ against the baselines across various layer configurations (full and lite) and embedding dimensions (32, 64, 128, 768), are presented in Table~\ref{tab:main_results}.

 The results consistently demonstrate the superiority of $\text{m}^3\text{BERT}$ framework. Across all evaluated configurations, both $\text{m}^3\text{BERT-s2}$ and $\text{m}^3\text{BERT-s3}$ significantly outperform the baseline models. 
This advantage is evident not only when using the full model depth and standard 768 embedding dimension but also extends robustly to the ``lite'' models (1/3 layers) and lower matryoshka embedding dimensions (32, 64, 128).

Particularly noteworthy is the performance of $\text{m}^3\text{BERT-s3}$. In our experiments, we observed that 
$\text{m}^3\text{BERT-s2}$ had already reached a state of convergence on the general-purpose multilingual corpora from Stage 2, and further pretraining on the same data yielded negligible gains. The continued pretraining on the large-scale web domain corpus (Stage 3) provides a substantial boost in retrieval effectiveness, showcasing the benefit of domain-specific adaptation combined with our multigranular pretraining strategy. Even at very low dimensions like 32 or 64, $\text{m}^3\text{BERT-s3}$ surpasses or performs competitively with baseline models using much larger embeddings, highlighting the efficiency and representational power instilled by our approach. The consistent gains across the spectrum of model sizes and embedding dimensions underscore the flexibility and strong generalization capabilities of $\text{m}^3\text{BERT}$, making it a highly effective foundation for industrial applications with varying resource constraints.

\begin{table}[htbp]
\centering
\caption{Recall@100 comparison of Matryoshka vs. Standard pretraining across datasets and embedding dimensions.}
\label{tab:mrl_vs_standard_pretrain}
\resizebox{0.47\textwidth}{!}{%
\begin{tabular}{@{}llcccc@{}}
\toprule
Dataset & Strategy & D=32 & D=64 & D=128 & D=768 \\
\midrule
\multirow{2}{*}{\textsc{Bing-Click}} 
    & Standard   & 57.19 & 62.50 & 64.95 & 65.55 \\
    & Matryoshka & 57.53 & 62.79 & 64.62 & 65.44 \\
\midrule
\multirow{2}{*}{\textsc{MS MARCO}} 
    & Standard   & 50.08 & 57.98 & 61.25 & 63.91 \\
    & Matryoshka & 51.55 & 58.63 & 61.29 & 63.28 \\
\midrule
\multirow{2}{*}{\makecell{\textsc{Natural}\\\textsc{Questions}}}
    & Standard   & 81.17 & 88.36 & 91.56 & 91.80 \\
    & Matryoshka & 83.74 & 88.49 & 91.09 & 91.32 \\
\midrule
\multirow{2}{*}{\textsc{TREC-COVID}} 
    & Standard   & 76.07 & 83.58 & 86.92 & 88.07 \\
    & Matryoshka & 79.23 & 84.14 & 86.33 & 87.36 \\
\bottomrule
\end{tabular}
}
\end{table}

\subsection{Impact of Matryoshka Pretraining}
\label{sec:matryoshka_pretrain_impact}
To assess the direct benefits of our multigranular matryoshka pretraining strategy, we compare it against standard pretraining approaches where models are individually optimized for specific layer depths and embedding dimensions.
We conduct two main pretraining paradigms for 100,000 steps using the Stage-1 English corpus, identical hyperparameters and architecture:

\begin{itemize}
    \item \textbf{Standard Pretraining}: For this paradigm, we train separate models. Each model is pretrained using a standard MLM objective where the loss is computed *only* from a specific target layer (the 4th) and for a specific target full-dimension (32, 64, 128, or 768). This results in 4 distinct pretrained models.
    \item \textbf{Matryoshka Pretraining}: Our proposed approach, where a single model is pretrained. The MLM loss is aggregated across multiple selected layers (including the 4th and 12th) and multiple sub-dimensions (32, 64, 128, and 768), as described in Section~\ref{sec:method_mrl}. 
\end{itemize}
All pretrained models adopt the $\text{m}^3\text{BERT}$ architecture and are independently fine-tuned (SFT) on each downstream dataset: \textsc{Bing-Click}, \textsc{MS MARCO}, \textsc{Natural Questions}, and \textsc{TREC-COVID}.
For all of the standard pretrained models, SFT targets the specific layer and dimension they were pretrained for. For the Matryoshka model, SFT also targets specific layer/dimension combinations for evaluation, but benefits from the joint pretraining.

As shown in Table~\ref{tab:mrl_vs_standard_pretrain}, the single matryoshka-pretrained model achieve comparable performance to individually pretrained counterparts. Notably, while performance at the highest dimension may show minor reductions in some datasets, matryoshka pretraining consistently improves retrieval quality at lower dimensions. This indicates that matryoshka pretraining enhances the flexibility of representation learning, particularly benefiting resource-constrained deployment scenarios where compact embeddings are crucial.

\subsection{Deployment}
\label{sec:deployment}
The model was mainstreamed in 2025 as a core component of the query–keyword online selection workflow, where it consistently handles 25,000 queries per second (QPS) under production traffic. To support this scale, we deployed a large-scale document index containing over 120 million keywords, enabling coverage across diverse query intents and verticals. Retrieval is powered by an approximate nearest neighbor (ANN) search framework, which efficiently narrows down the search space and returns the top 100 most relevant keywords for each incoming query with low latency.
From a business perspective, this deployment has proven to be highly impactful. By integrating the model directly into the serving stack, we improved both the efficiency and precision of query–keyword matching, driving measurable gains in downstream ad relevance and monetization. The system currently contributes an annualized revenue impact (APR) of approximately USD 50 million, establishing it as one of the most valuable model deployments in the query understanding pipeline.

\section{Analysis}
\label{sec:analysis}
In this section, we analyze the contributions of different components of $\text{m}^3\text{BERT}$ and explore more matryoshka strategies. 
Section~\ref{sec:architecture_ablation} shows the impact of various modern architectural components on model performance. Section~\ref{sec:matryoshka_distillation} explores how leveraging higher-dimensional representations as teacher signals can enhance the performance of lower-dimensional embeddings during matryoshka pretraining. Section~\ref{sec:matryoshka_finetuning} demonstrates the advantages of using models pretrained with matryoshka strategy in conjunction with matryoshka supervised fine-tuning.

\begin{table*}[h]
\centering
\caption{Matryoshka Supervised Finetuning (MRL SFT) performance on \textsc{Bing-Click} (Lite models). Parenthesized values denote the performance drop ($\Delta$) compared to standard single-dimension finetuning. \textbf{Bold} indicates the smallest performance drop in each column.}
\label{tab:mrl_finetuning_comparison}
\resizebox{\textwidth}{!}{
\begin{tabular}{@{}lcccccccc@{}}
\toprule
\multirow{3.5}{*}{Model (Lite)} & \multicolumn{2}{c}{Dim=32} & \multicolumn{2}{c}{Dim=64} & \multicolumn{2}{c}{Dim=128} & \multicolumn{2}{c}{Dim=768} \\
\cmidrule(lr){2-3} \cmidrule(lr){4-5} \cmidrule(lr){6-7} \cmidrule(lr){8-9}
                       & R@100 & R@1000 & R@100 & R@1000 & R@100 & R@1000 & R@100 & R@1000 \\
\midrule
mBERT             & 57.12 (-0.34) & 80.66 (-0.12) & 62.29 (-0.21) & 84.59 (-0.08) & 63.87 (-0.71) & 85.87 (-0.40) & 64.26 (-1.13) & 86.20 (-0.79) \\
mE5               & 56.91 (-0.27) & 80.33 (-0.30) & 61.87 (-0.29) & 84.20 (-0.20) & 63.44 (-0.96) & 85.42 (-0.72) & 63.99 (-1.40) & 85.91 (-0.99) \\
ModernBERT        & 56.34 (-0.76) & 80.24 (-0.31) & 62.29 (-0.14) & 84.65 \textbf{(+0.06)} & 63.80 (-0.79) & 85.79 (-0.44) & 64.11 (-1.23) & 86.00 (-0.87) \\
$\text{m}^3\text{BERT-s2}$ & 58.25 \textbf{(-0.09)} & 81.59 \textbf{(-0.08)} & 63.52 \textbf{(+0.09)} & 85.51 (+0.02) & 65.01 (-0.46) & 86.64 (-0.38) & 65.36 \textbf{(-0.97)} & 86.96 \textbf{(-0.70)} \\
$\text{m}^3\text{BERT-s3}$  & 61.73 (-0.33) & 84.30 (-0.22) &   66.45 (-0.02) & 87.56 (-0.11) & 67.39 \textbf{(-0.33)} & 88.19 \textbf{(-0.32)} & -- & -- \\
\bottomrule
\end{tabular}%
}
\end{table*}

\begin{table}[ht]
\centering
\caption{Architecture ablation for $\text{m}^3\text{BERT}$ on \textsc{Bing-Click}.}
\label{tab:arch_ablation}
\begin{tabular}{lcc}
\toprule
Model Configuration & R@100 & R@1000 \\
\midrule
$\text{m}^3\text{BERT}$ &  61.65 &  84.13 \\
\quad -- SwiGLU &  60.18 &  82.79 \\
\quad -- Pre-norm &  61.42 &  84.01 \\
\quad -- RMSNorm &  61.66 &  84.12 \\
\quad + Hidden Dropout &  61.65 &  84.14 \\
\quad + Bias Terms &  61.53 &  84.12 \\
\bottomrule
\end{tabular}
\end{table}

\subsection{Architecture Ablation}
\label{sec:architecture_ablation}
To quantify the impact of the modern architectural modifications incorporated into $\text{m}^3\text{BERT}$, we perform an ablation study. Starting with our $\text{m}^3\text{BERT-s2}$ model in a lite configuration (1/3 layers) and using 64-dimensional embeddings, we systematically revert each enhancement to its counterpart in traditional BERT or E5 models. Specifically, we evaluate the following variations:
\begin{itemize}
    \item Replacing SwiGLU with GeLU activation (\textit{w/o SwiGLU}).
    \item Reverting from pre-normalization to post-normalization (\textit{w/o Pre-norm}).
    \item Replacing RMSNorm with standard LayerNorm (\textit{w/o RMSNorm}).
    \item Re-introducing dropout on hidden states between transformer layers (\textit{w/ Dropout}).
    \item Re-introducing the bias terms in self-attention and feed-forward layers (\textit{w/ Bias}).
\end{itemize}
Each ablated model is trained on the \textsc{Bing-Click} dataset without pretraining. The results, presented in Table~\ref{tab:arch_ablation}, show that while some architectural choices positively impact the final performance, others exhibit negligible effects on accuracy but, as noted by ~\citep{llama}, contribute to computational efficiency or training stability, justifying their inclusion in $\text{m}^3\text{BERT}$.

\subsection{Matryoshka Distillation}
\label{sec:matryoshka_distillation}
Given the nature of matryoshka pretraining, which simultaneously optimizes embeddings across multiple layers and dimensions, we explore leveraging deeper, higher-dimensional embeddings as "teacher" signals for shallower, lower-dimensional embeddings during the pretraining process itself. This can be viewed as a form of self-distillation inherent to the matryoshka framework. To investigate this, we take the $\text{m}^3\text{BERT}$ model checkpoint from Stage 1 after 300,000 pretraining steps as our base. We then continue pretraining for an additional 50,000 steps under several distillation settings. The distillation objective aims to minimize the KL divergence between the output distributions of the LM heads corresponding to the teacher and student embeddings.

The specific settings are as follows:
\begin{itemize}
    \item \textbf{Continued MRL Pretraining}: Continue standard matryoshka pretraining for 50k steps (baseline for this experiment).
    \item \textbf{Distill L12-D768 to All}: The embedding from the 12th layer and 768 dimensions acts as the teacher. Its LM head output is used to distill knowledge to the LM head outputs of all 11 other combinations of layers (4th, 8th, 12th) and dimensions (32, 64, 128, 768), excluding the teacher itself.
    \item \textbf{Specific Distillation}: Use the LM head output from the 12th layer as the teacher and distill to the 4th layer, 64-dimension embedding. We experiment with two variants: (1) teacher = L12-D64, and (2) teacher = L12-D768.
\end{itemize}
We evaluate the performance of the 4th layer, 64-dimension embedding (L4-D64) after these continued pretraining variants on the \textsc{Bing-Click} dataset. The results in Table~\ref{tab:matryoshka_distillation} suggest that leveraging deeper or higher-dimensional embeddings as teacher signals during pretraining can indeed enhance the performance of shallower, lower-dimensional student embeddings. However, the "Distill L12-D768 to All" setting, where a single powerful teacher attempts to guide numerous diverse students simultaneously, does not show a comparable benefit
. This might indicate that overly broad distillation targets could introduce conflicting signals or noise, potentially interfering with the nuanced learning required for each specific layer-dimension combination.

\begin{table}[ht]
\centering
\caption{Impact of Matryoshka Distillation.
}
\label{tab:matryoshka_distillation}
\begin{tabular}{@{}lcc@{}}
\toprule
Pretraining Strategy & R@100 & R@1000 \\
\midrule
MRL Pretraining (300k steps) & 62.96 & 85.15 \\
\midrule
\multicolumn{3}{@{}l}{\textit{+ 50k additional steps}} \\
\quad MRL Pretraining        & 63.00 & 85.20 \\
\quad Distill L12-D768 to All & 62.98 & 85.20 \\
\quad Distill L12-D64 to L4-D64      & 63.12 & \textbf{85.30} \\
\quad Distill L12-D768 to L4-D64     & \textbf{63.14} & 85.27 \\
\bottomrule
\end{tabular}
\end{table}

\subsection{Matryoshka Finetuning}
\label{sec:matryoshka_finetuning}

We also investigate the application of the MRL principle during the supervised fine-tuning (SFT) phase. Standard SFT typically optimizes a model for a single embedding dimension, which often yields the best possible performance at that specific dimension. MRL SFT aims to make the model perform well across a set of chosen dimensions ($\mathcal{D}_{\text{SFT}} = \{32, 64, 128, 768\}$ for $\text{m}^3\text{BERT-s2}$, and $\mathcal{D}_{\text{SFT}} = \{32, 64, 128\}$ for $\text{m}^3\text{BERT-s3}$), aggregating SFT loss contributions from these dimensions. This explicit optimization for multiple granularities can sometimes introduce a 
slight performance decrease at any single dimension compared to dedicated standard SFT.

Table~\ref{tab:mrl_finetuning_comparison} presents the performance of Lite models fine-tuned with the MRL SFT strategy, alongside the performance change (in parentheses) relative to their standard SFT counterparts (from Table~\ref{tab:main_results}, Lite section).
While MRL SFT can lead to minor performance drops across all models, our $\text{m}^3\text{BERT}$ models
, having been pretrained with an MRL objective, 
generally exhibit smaller performance degradation compared to the baseline models when subjected to MRL SFT.
This indicates that MRL pretraining effectively prepares the model architecture for subsequent MRL SFT, allowing it to maintain strong, flexible performance across various embedding dimensions with a more mitigated trade-off.

\section{Related Work}

\subsection{Pretrained Embedding Models}

The advent of encoder-only Transformer architectures, exemplified by BERT~\citep{bert} and its successors like RoBERTa~\citep{roberta}, has significantly advanced information retrieval by providing powerful contextualized text embeddings. Subsequent specialized models such as Sentence-BERT~\citep{reimers-2019-sentence-bert}, and more recently E5~\citep{e5} and ModernBERT~\citep{modernbert}, have further refined these embeddings for retrieval tasks through techniques like contrastive learning~\citep{gao2021simcse, reimers-2019-sentence-bert, lan2025llave} and instruction tuning~\citep{e5, wei2021finetuned}. These models are typically pretrained on vast corpora and then fine-tuned for specific domains.

However, a critical limitation of these existing pretrained embedding models is their fixed architecture and output dimension. This inherent rigidity makes it challenging to efficiently adapt them to diverse downstream applications with varying latency and computational resource constraints, common in commercial systems. While solutions like task-specific fine-tuning of smaller models by partial initialization or post-hoc compression~\citep{hinton2015distilling, han2015deep, jacob2018quantization} exist, they often fail to fully leverage the extensive knowledge of large pretrained models.

\subsection{Matryoshka Representation Learning}

Matryoshka Representation Learning (MRL)\citep{mrl} presents a novel approach to create adaptable representations by training a single model
whose embeddings are effective even when truncated to various smaller nested dimensions
. This allows for a direct trade-off between performance and computational cost without needing to train or store multiple distinct models, and has shown promise in vision\citep{mrl-recommendation-v} and language tasks~\citep{mrl-recommendation} by enabling flexible deployment. Beyond output embeddings, the core principle of MRL has also been extended to internal model components~\citep{matformer, wang2025training}, and its versatility is further demonstrated in multimodal learning~\citep{matrymulti, matryvlm} and federated learning for knowledge sharing~\citep{federated}.

While MRL offers valuable dimensional flexibility, its application
focused on the output embeddings, often during fine-tuning stages~\citep{mrl-recommendation}, rather than being integrated into the pretraining of foundation models. Furthermore, existing MRL work primarily addresses variability in embedding dimension~\citep{mrl}, not 
extending this adaptive capability to the model's architectural depth. Our work is inspired by MRL's efficiency but aims to extend its principles by incorporating multigranular optimization across both embedding dimensions and transformer layers directly within the pretraining phase, creating a more fundamentally versatile foundation model.

\subsection{Efficient Model Deployment}
To bridge the gap between large, powerful models and resource-constrained deployment environments, a variety of techniques have gained prominence. Knowledge distillation~\citep{hinton2015distilling} focuses on transferring complex capabilities from teacher to student models~\citep{hsieh2023distilling, miao2023exploring, gu2024minillmknowledge}. Other methods operate directly on a trained network, such as pruning techniques which can remove significant weights in a one-shot manner~\citep{frantar2023sparse, sun2024simple}, and advanced quantization methods that drastically reduce the numerical precision of parameters with minimal performance loss~\citep{dettmers2023qlora, frantar2023gptqa}. A different approach involves dynamic networks that adapt computation at inference time through adaptive layer skipping or early exiting based on input confidence~\citep{schuster2022confident}.

These methods have limitations in the context of flexible, large-scale deployment. Knowledge distillation requires separate training and maintaining a student model for each specific performance target, which can be computationally expensive and add significant engineering overhead. Dynamic methods like early exiting 
are not directly controllable for deploying a model variant tailored to a specific hardware profile
. Pruning and quantization are largely orthogonal to our work and can be combined with our approach. 

\section{Conclusion}
We proposed $\text{m}^3\text{BERT}$, a Modern, Multilingual, Matryoshka Bidirectional Encoder that leverages a novel matryoshka pretraining strategy to optimize embedding representations across multiple transformer layers and dimensions. Extensive experiments on the BINGCLICK dataset demonstrate that $\text{m}^3\text{BERT}$ consistently outperforms state-of-the-art models across various model sizes and embedding dimensions, effectively balancing resource efficiency and retrieval performance in industrial settings. 
The model's effectiveness is also verified by its successful deployment at Bing.

\clearpage
\bibliographystyle{ACM-Reference-Format}
\bibliography{sample-base}

\appendix

\section{Implementation Details}
\subsection{Evaluation}
As mentioned in Section~4.1, we conduct experiments on four benchmark datasets: BING-CLICK, MS MARCO Document Ranking, Natural Questions, and TREC-COVID. All datasets are formatted as query-document pairs to align with our model's input requirements. Specifically, for the Natural Questions dataset, we use question-answer pairs where the answer span is treated as the relevant document. For TREC-COVID, we adopt the title-text pairs, where the title represents the query and the document text serves as the candidate passage.

Our evaluation pipeline leverages the sentence-transformers for generating text embeddings from the fine-tuned models.
For the retrieval tasks, we employ Faiss, a library for efficient similarity search. Specifically, we perform an exhaustive (exact) search using the inner product similarity (cosine similarity on L2-normalized embeddings). This is typically achieved using Faiss indices like \texttt{IndexFlatIP} after ensuring all embeddings are L2-normalized. No approximate nearest neighbor (ANN) search methods were used for the reported results to ensure precise recall figures.

The Recall@K metric is computed as follows:
For each query $q$ in the test set and its corresponding ground-truth positive document $d^+ \in \mathcal{D}_{\text{pool}}$ (where $\mathcal{D}_{\text{pool}}$ is the 10 million document candidate pool):
\begin{enumerate}
    \item Obtain the query embedding $E(q)$ and the embeddings for all documents $d_i \in \mathcal{D}_{\text{pool}}$, denoted $E(d_i)$.
    \item Calculate the similarity score (cosine similarity) between $E(q)$ and every $E(d_i)$.
    \item Rank all documents in $\mathcal{D}_{\text{pool}}$ based on their similarity scores with $E(q)$ in descending order.
    \item If the ground-truth positive document $d^+$ is found within the top K ranked documents, this query is considered a "hit" for Recall@K.
    \item The final Recall@K is the proportion of queries for which $d^+$ was retrieved within the top K results, averaged over all test queries.
\end{enumerate}
We report Recall@100 and Recall@1000 in our experiments.

\subsection{Matryoshka Distillation}
As explored in Section 4.3.3, Matryoshka distillation aims to leverage deeper or higher-dimensional embeddings as teacher signals for shallower or lower-dimensional student embeddings during pretraining. The distillation objective minimizes the Kullback-Leibler (KL) divergence between the predicted token distributions from the LM heads of the teacher and student embeddings.
Given a teacher embedding output $h_T$ (e.g., from layer $l_T$, dimension $d_T$) and a student embedding output $h_S$ (e.g., from layer $l_S$, dimension $d_S$), their respective predicted token distributions over the vocabulary $V$ are $P_T = \text{Softmax}(h_T W_{[:d_T,:]} / \tau_D)$ and $P_S = \text{Softmax}(h_S W_{[:d_S,:]} / \tau_D)$, where $W$ is the shared MLM projection matrix and $\tau_D$ is the distillation temperature. The distillation loss for a single teacher-student pair is:
\begin{equation}
    \mathcal{L}_{\text{distill}}(P_S || P_T) = \sum_{v \in V} P_S(v) \log \frac{P_S(v)}{P_T(v)}
\end{equation}
When multiple student (or teacher) embeddings are involved, such as in the "Distill L12-D768 to All" setting, the individual KL divergence losses are summed. This distillation loss is then added to the primary Matryoshka pretraining loss (e.g., multigranular MLM loss $\mathcal{L}_{\text{MRL}}$):
\begin{equation}
    \mathcal{L}_{\text{total}} = \mathcal{L}_{\text{MRL}} + \lambda_D \sum_{(S,T) \in \mathcal{P}_{\text{distill}}} \mathcal{L}_{\text{distill}}(P_S || P_T)
\end{equation}
where $\mathcal{P}_{\text{distill}}$ is the set of chosen teacher-student pairs and $\lambda_D$ is a weighting factor for the distillation loss (set to $1.0$ in our experiments).

We observed that initiating Matryoshka Distillation from the very beginning of pretraining can lead to instability. This is likely because the initial representations, especially from deeper layers and higher dimensions, are noisy and not yet effective teachers. Therefore, in our experiments reported in Table~\ref{tab:matryoshka_distillation}, we first pretrained the $\text{m}^3\text{BERT}$ model for 300,000 steps using standard Matryoshka pretraining (as in Stage 1) and then continued pretraining for an additional 50,000 steps with the distillation objective incorporated. The distillation temperature $\tau_D$ was set to $1.0$. Other hyperparameters (learning rate, batch size, etc.) for these 50,000 steps remained consistent with the Stage 1 pretraining configuration. This approach shows promise for enhancing the quality of more compact representations within the Matryoshka framework.

\subsection{Matryoshka Finetuning}
For Matryoshka Finetuning, as discussed in Section 4.3.4, we adapt the standard supervised fine-tuning (SFT) objective  to simultaneously optimize the model for multiple embedding dimensions.
Given a set of target embedding dimensions $\mathcal{D}_{\text{SFT}}$ (e.g., $\{32, 64, 128, 768\}$ for $\text{m}^3\text{BERT-s2}$ and $\{32, 64, 128\}$ for $\text{m}^3\text{BERT-s3}$), the MRL SFT loss is the sum of the standard SFT losses computed for each dimension in $\mathcal{D}_{\text{SFT}}$:
\begin{equation}
    \mathcal{L}_{\text{MRL-SFT}} = \sum_{d \in \mathcal{D}_{\text{SFT}}} \mathcal{L}_{\text{SFT}}(E_{[:d]})
\end{equation}
where $E_{[:d]}$ denotes the model's output embedding truncated to the first $d$ dimensions, and $\mathcal{L}_{\text{SFT}}(E_{[:d]})$ is the contrastive loss calculated using these $d$-dimensional embeddings. The individual SFT losses for different dimensions are directly summed.
All other hyperparameters for MRL SFT (e.g., learning rate, batch size, temperature $\tau$ for the contrastive loss) were kept identical to those used for the standard single-dimension SFT experiments reported in the main results.

\section{Pretraining Data}
For all of the pretraining stages, input sequences were processed by padding shorter sequences and truncating longer ones to a maximum length of 1024 tokens using mE5 tokenizer.

\paragraph{Nemotron-CC (Stage 1)}
The English pretraining corpus is derived from the high-quality split of Nemotron-CC, which is itself based on CommonCrawl (CC) and refined through extensive filtering and deduplication. We utilized a subset of approximately 100 billion tokens from the high-quality split.

\begin{table*}[htbp]
\centering
\caption{Smoothed sampling probabilities for the top 20 languages (Stage 2 multilingual Wikipedia corpus).}
\label{tab:multilingual_distribution_twocol}
\begin{tabular}{@{}cl r @{\hspace{2em}} cl r@{}} 
\toprule
Code & Language Name & Prob.(\%) & Code & Language Name & Prob.(\%) \\
\cmidrule(lr){1-3} \cmidrule(lr){4-6} 
en  & English         & 20.99 & ceb & Cebuano         & 20.33 \\
de  & German          & 11.89 & sv  & Swedish         & 11.08 \\
fr  & French          & 11.05 & nl  & Dutch           & 9.73  \\
ru  & Russian         & 9.11  & es  & Spanish         & 8.77  \\
it  & Italian         & 8.74  & arz & Egyptian Arabic & 8.02  \\
pl  & Polish          & 7.90  & ja  & Japanese        & 7.20  \\
zh  & Chinese         & 7.18  & uk  & Ukrainian       & 6.85  \\
vi  & Vietnamese      & 6.83  & war & Waray-Waray     & 6.75  \\
ar  & Arabic          & 6.57  & pt  & Portuguese      & 6.16  \\
fa  & Persian         & 5.64  & ca  & Catalan         & 4.62  \\
\bottomrule
\end{tabular}%
\end{table*}

\paragraph{Multilingual Wikipedia (Stage 2)}
\label{app:multilingual_wiki}
For multilingual adaptation, we used publicly available Wikipedia dumps \footnote{\url{https://meta.wikimedia.org/wiki/List_of_Wikipedias}} for the top 100 languages by article count. The selection of languages and initial processing followed mBERT. The raw data from these dumps, totaling approximately 20 billion tokens, was further processed. As mentioned in the main text, an exponential smoothing strategy ($P'(L) \propto P(L)^S$ with $S=0.7$) was applied to upsample low-resource languages and downsample high-resource languages. This creates a more balanced multilingual training mixture, preventing high-resource languages from dominating the training process.
Table~\ref{tab:multilingual_distribution_twocol} illustrates the smoothed sampling probabilities for the top 20 languages in our multilingual corpus after applying this strategy.

\paragraph{Web Domain Corpus (Stage 3)}
\label{app:ads_corpus}
The corpus for Stage 3 continual pretraining, designed to specialize the model for ads and web search-related industrial retrieval tasks, consists of 10 billion query-item pairs derived from anonymized and aggregated user interaction logs within an advertising system. These pairs represent implicit positive associations between user queries and relevant advertising content. We constructed three primary types of query-item pairs:
\begin{itemize}
    \item \textbf{(Query, Keyword):} Pairs linking user search queries to advertiser-defined keywords that triggered ad impressions.
    \item \textbf{(Query, Ad Creative):} Pairs linking user search queries to the textual content of advertisements they interacted with. The "Ad Creative" item concatenates the ad's title and descriptive text.
    \item \textbf{(Query, Landing Page Information):} Pairs linking user search queries to textual information associated with the ad's destination. The "Landing Page Information" item is a concatenation of text extracted from resources related to the ad's landing URL, such as its associated page title, prominent headings, and representative content snippets. 
 Compared to the carefully curated and processed \textsc{Bing-Click} evaluation data, this pretraining corpus for Stage 3 comprises a significantly larger volume of text that is more diverse in form, contains more noise, and is generally of lower raw quality, reflecting the scale and nature of real-world advertising interaction data.
\end{itemize}
These different pair types capture diverse semantic relationships relevant to advertising retrieval. In the underlying advertising ecosystem, a single keyword can be associated with multiple ad creatives, and an ad creative can point to various landing page information items (and vice versa). This rich interconnectedness provides a diverse training signal.

The corpus was rigorously pre-processed, including the duplication of identical query-item pairs and query-level sampling caps to mitigate frequency bias. The model was trained on this corpus for approximately 2 trillion tokens using the Infinite Contrastive Learning (Inf-CL). The input for the "item" side of the pair was formed by concatenating the relevant textual fields (e.g., ad title + ad description for "Ad Creative") using special separator tokens, then tokenized and processed to the model's maximum input length. This domain-specific pretraining significantly enhances the model's utility for commercial retrieval applications.

\section{Use of Large Language Models}
\label{sec:llm_usage}

In preparing this manuscript, the LLM was used for language polishing, such as rephrasing sentences for clarity and correcting grammar. The human authors critically reviewed and edited all LLM-generated outputs, and retain full responsibility for the final content, methodology, and conclusions of this work.

\end{document}